\documentclass[conference]{IEEEtran}
\IEEEoverridecommandlockouts
\usepackage{cite}
\usepackage{bm}
\usepackage{amsmath,amssymb,amsfonts}
\usepackage{graphicx}
\usepackage{multirow}
\usepackage{textcomp}
\usepackage{soul}
\usepackage{multicol,xcolor,color,colortbl,hhline,caption}
\usepackage[linesnumbered,ruled]{algorithm2e}
\def\BibTeX{{\rm B\kern-.05em{\sc i\kern-.025em b}\kern-.08em
    T\kern-.1667em\lower.7ex\hbox{E}\kern-.125emX}}
\usepackage[bookmarks=false]{hyperref}

\begin{document}

\title{$\text{O}^2$PF: Oversampling via Optimum-Path Forest for Breast Cancer Detection
\thanks{$^\ast$Authors contributed equally.}
\thanks{The authors would like to thank FAPESP grants \#2013/07375-0, \#2014/12236-1, \#2019/18287-0, and \#2019/07665-4, as well as CNPq grants \#427968/2018-6 and \#307066/2017-7.}
}		

\author{\IEEEauthorblockN{Leandro A. Passos$^\ast$, Danilo S. Jodas$^\ast$, Luiz C. F. Ribeiro$^\ast$, Thierry Pinheiro, Jo\~ao P. Papa}
\IEEEauthorblockA{Department of Computing \\
S\~ao Paulo State University\\
Bauru, Brazil \\
\{leandro.passos, luiz.felix, joao.papa\}@unesp.br, \{danilojodas, thierrypin\}@gmail.com}}

\maketitle

\begin{abstract}
Breast cancer is among the most deadly diseases, distressing mostly women worldwide. Although traditional methods for detection have presented themselves as valid for the task, they still commonly present low accuracies and demand considerable time and effort from professionals. Therefore, a computer-aided diagnosis (CAD) system capable of providing early detection becomes hugely desirable. In the last decade, machine learning-based techniques have been of paramount importance in this context, since they are capable of extracting essential information from data and reasoning about it. However, such approaches still suffer from imbalanced data, specifically on medical issues, where the number of healthy people samples is, in general, considerably higher than the number of patients. Therefore this paper proposes the $\text{O}^2$PF, a data oversampling method based on the unsupervised Optimum-Path Forest Algorithm. Experiments conducted over the full oversampling scenario state the robustness of the model, which is compared against three well-established oversampling methods considering three breast cancer and three general-purpose tasks for medical issues datasets.
\end{abstract}


\section{Introduction}
\label{s.introduction}

Computer-Aided Diagnosis (CAD) systems aim at helping physicians to quickly report better diagnosis to patients, thus representing an essential step towards dangerous disease accurate diagnosis. Besides, intelligent-based CAD systems have successfully employed machine learning techniques, a promising subfield of artificial intelligence, to tackle complicated problems that demand knowledge and reasoning about the subject. Regarding the latter, several works were developed in the last few years to aid detecting atherosclerosis~\cite{JodasESWA:2016}, Parkinson's disease~\cite{ribeiro2019bag}, and breast cancer~\cite{YassinCMPB:2018}, to name a few.

Breast cancer is a dangerous illness, affecting millions of women, as well as few men, around the world. According to the World Health Organization, such a disease was responsible for the death of approximately $627,000$ women in $2018$ worldwide~\cite{whowebsite:2020}. Therefore, early diagnosis is crucial to effectively prevent its progress, consequently making it possible to elaborate more efficient treatment plans. In general, standard breast cancer diagnosis consists of a visual analysis performed by professionals aiming to identify possible abnormalities (e.g., nodules), which may indicate cancer risk signs. Once identified, it is possible to extract relevant measures from such nodules, assisting physicians to judge the presence or absence of cancerous tissue.

Many image processing-based algorithms have also been proposed to help in such tasks~\cite{XianPR:2018, WieclawekITB:2019, LiuMBEC:2017}, as well as machine learning techniques, which have been commonly employed for both nodule segmentation and classification. Agarap~\cite{agarap2018breast}, for instance, studied the performance of several machine learning techniques to label a nodule as benign or malignant. Further, Passos et al.~\cite{passosECCOMAS:19} developed a neural network to tell benign from malignant nodules, and to label the latter according to the most likely cancer type. 

Moreover, once the nodules can be seen as abnormalities, their identification can also be tackled in an unsupervised fashion. In such a direction, Ribeiro et al.~\cite{ribeiroISCBMS:2015} used the unsupervised Optimum-Path Forest (OPF)~\cite{RochaIJIST:09} algorithm to distinct malignant nodes from benign, obtaining significant results. The unsupervised Optimum-Path Forest algorithm possesses properties commonly employed to improve other machine learning techniques, such as the Brainstorm Optimization~\cite{afonsoSACI:2018}, as well as to create new techniques, such as the OPF-based approach for anomaly detection~\cite{guimaraesIEEENETWORK:2018} and the Fuzzy OPF classifier~\cite{SouzaIEEETFS:19}.

Although the above-mentioned techniques obtained promising results in the context of breast cancer detection, most of the classification algorithms usually suffer from imbalanced dataset problems, which arises when the number of samples among classes differs significantly. In this scenario, the trained classifier is more likely to label a new sample as belonging to the most common (majority) class, degrading its performance for the smallest (minority) class. Moreover, such a situation may degenerate when the dataset presents outliers. Numerous studies have also been proposed to cope with such a problem in medical datasets~\cite{ZuIEEEAccess:2018, RazzaghiAOR:2019, RezaeiMTAA:2019}. Therefore, the generation of synthetic samples for the minority class, which is usually referred to as oversampling, is recognized as a prominent contribution to rebalancing the dataset for classifier training purposes, consequently improving their robustness to label minority class samples correctly. In this context, several powerful oversampling strategies have been proposed in the literature to tackle such a problem~\cite{ChawlaJAIR:02,han:05borderline,He:2008adasyn}, most of them still presenting difficulties while enforcing diversity among new synthetic samples, which denotes a problem that worth to be addressed, since it can improve classifier generalization properties.

To such an extent, this work proposes an approach to perform oversampling via the Optimum-Path Forest, hereinafter named $\text{O}^2$PF. The method employs the unsupervised OPF algorithm to capture features intrinsic to the minority class into different clusters. Further, new training examples are generated by sampling from a Gaussian distribution parametrized by the cluster characteristics. Therefore, the main contribution of this paper are twofold: (i)~extending the OPF algorithm capabilities by introducing a novel oversampling mechanism that enforces synthetic intra-class variability; and (ii)~studying how $\text{O}^2$PF can benefit the development of CAD systems by extensively evaluating it on five tumor-classification and one retinopathy identification tasks. Finally, the full-balanced datasets are employed to train the OPF classifier, and results, i.e., the accuracy, recall, and F1-measure, are compared against the ones obtained by the standard training dataset without oversampling. Additionally, results are compared against three baselines: SMOTE~\cite{ChawlaJAIR:02}, the Borderline SMOTE~\cite{han:05borderline}, and ADASYN~\cite{He:2008adasyn}. 

In the remainder of this paper, Section~\ref{s.background} presents the theoretical background regarding the unsupervised and supervised OPF algorithm variants, whereas Section~\ref{s.proposed} introduces the proposed approach. Further, Section~\ref{s.methodology} outlines the experimental setup and Section~\ref{s.experiments} discusses experimental results. Finally, Section~\ref{s.conclusion} presents conclusions and future works.

\section{Background}
\label{s.background}

This section presents the theoretical background regarding the unsupervised and supervised variants of the OPF algorithm.

\subsection{Unsupervised Optimum-Path Forest}
\label{ss.unsupervised_opf}

Let $\mathcal{D} = \left\{ \bm{x}_1, \bm{x}_2, \ldots, \bm{x}_n \right\}$ be a dataset such that $\bm{x}_i \in \mathbb{R}^m$ represents the features extracted from the $i$-th sample. Further, let $\mathcal{G}=(\mathcal{D}, \mathcal{A}_{k^\star})$ be a graph where each node corresponds to a different feature vector connected to its $k^\star$-nearest neighbors, as defined in the adjacency relationship set $\mathcal{A}_{k^\star}$.

The unsupervised OPF algorithm consists in partitioning the graph through a competitive process, in which a few samples are marked as ``prototypes'' and compete among themselves to conquer the remaining nodes. Such a procedure partitions the $\mathcal{G}$ into optimum-path trees (OPTs) rooted at a prototype, each corresponding to a cluster, where a sample is more similar to the elements of its tree than any other tree. Overall, the process can be divided into three steps: (i) computing a proper neighborhood size $k^\star$ and the adjacency relationship $\mathcal{A}_{k^\star}$, (ii) electing the prototype nodes, and (iii) performing the competition process to partition the dataset into OPTs.

Regarding the first step, different approaches may be considered. Among others, Rocha et al.~\cite{RochaIJIST:09} proposed to find $k^\star$ by minimizing the normalized graph cut function, as it takes into account the dissimilarities between clusters as well as the similarity degree among samples of each cluster.

Concerning the second step, the algorithm must select the prototypes to form the root of each OPT (which will ultimately form clusters) to rule the competition process and conquer the remaining samples in the graph. The supervised OPF proposed by Papa et al.~\cite{PapaIJIST:09, papa2012efficient} selects as prototypes the nearest samples from different classes, found by computing the graph Minimum Spanning Tree (MST). However, in the unsupervised variant, since labels are usually unavailable, Rocha et al.~\cite{RochaIJIST:09} proposed to select the prototypes as the samples located in the center of each cluster. To such an extent, all samples are assigned a density score $\rho(\bm{x}_i), \;\; \forall \bm{x}_i \in \mathcal{D}$, computed through a Gaussian probability density function (pdf), defined as follows:

\begin{equation}
    \rho(\bm{x}_i) = 
        \frac{1}{k^\star \sqrt{2\pi\sigma^2}}
        \sum_{\forall \bm{x}_j\in \mathcal{A}_{k^\star}(\bm{x}_i)}
        \exp \left(
            \frac{-d(\bm{x}_i,\bm{x}_j)}{2\sigma^2}
        \right),
    \label{e.pdf}
\end{equation}

\noindent
where $i \neq j$, $\sigma=d_{\max}/3$ and $d_{\max}$ stands for the maximum arc-weight in $\mathcal{G}$. This formulation considers all adjacent nodes for density computations, as the Gaussian distribution covers $99.7\%$ of the samples with distance $d(\bm{x}_i,\bm{x}_j) \in [0,3\sigma]$.

After evaluating Equation~\ref{e.pdf} for each node in the graph, the density values are used to populate a priority queue in a way that the unsupervised OPF maximizes the cost of each sample, thus partitioning the graph into OPTs. Such a cost is defined in terms of paths on $\mathcal{G}$, which is an acyclic sequence of adjacent samples in $\mathcal{A}_{k^\star}$. 

Let $\pi_{\bm{x}_i}$ be a path with terminus at sample $\bm{x}_i$ and starting from some root $\mathcal{R}(\bm{x}_j)$, being the latter the set of all prototype samples. Further, let $\pi_{\bm{x}_i} = \langle \bm{x}_i \rangle$ be a trivial path (i.e., a path containing only one sample), whereas $\pi_{\bm{x}_i} \cdot \langle \bm{x}_i,\bm{x}_j\rangle$ denotes the concatenation of a path $\pi_{\bm{x}_i}$ and the arc $(\bm{x}_i,\bm{x}_j)$ such that $i\neq j$.

In the third step, the algorithm assigns to each path $\pi_{\bm{x}_i}$ a value $f_{\min}(\pi_{\bm{x}_i})$ given by a smooth connectivity function $f_{\min}:{\mathcal{D}}\rightarrow \mathbb{R}^+$, which must satisfy some constraints to ensure the algorithm theoretic correctness~\cite{FalcaoIEEEPAMI:04, ciesielski2018smooth}. A path $\pi_{\bm{x}_i}$ is considered optimal if ${f_{\min}(\pi_{\bm{x}_i})\geq f_{\min}(\tau_{\bm{x}_i})}$ for any other path $\tau_{\bm{x}_i}$. Among the proposed path-cost functions in the literature, the unsupervised OPF relies on the following formulation:

\begin{eqnarray}
    \label{e.f_min}
    f_{\min}(\langle \bm{x}_i \rangle) & = &\left\{ \begin{array}{ll} 
        \rho(\bm{x}_i)           & \mbox{if $\bm{x}_i \in \mathcal{R}$} \\
        \rho(\bm{x}_i) - \delta   & \mbox{otherwise.} \\
    \end{array}\right.\\ \nonumber
    f_{\min}(\pi_{\bm{x}_i}\cdot \langle\bm{x}_i, \bm{x}_j\rangle) & = & \min \{f_{\min}(\pi_{\bm{x}_i}), \rho(\bm{x}_j)\},
\end{eqnarray}

\noindent
where $\delta = \min_{\forall (\bm{x}_i,\bm{x}_j)\in \mathcal{A}_{k^\star} | \rho(t) \neq \rho(s) } |\rho(t)-\rho(s)|$ consists in the smallest quantity to avoid plateaus and over-segmentation in regions near prototypes (i.e., areas with the highest density). 

Among all possible paths $\pi_{\bm{x}_i}$ that originate in some local maximum, i.e., some prototype, the OPF algorithm assigns to each sample a final path whose minimum density value along it is maximum. Such final path value is represented by a cost map $\mathcal{C}$, as follows: 

\begin{equation}
\label{e.min_cost_map}
    \mathcal{C}(\bm{x}_i) =
        \max_{\forall \pi_{\bm{x}_j} \in \mathcal{D}, \; i \neq j}
        \{
            f(\pi_{\bm{x}_j} \cdot \langle \bm{x}_j, \bm{x}_i \rangle)
        \}.
\end{equation}

The OPF algorithm maximizes $\mathcal{C}(\bm{x}_i) \; \forall \bm{x}_i \in \mathcal{D}$ by computing an optimum-path forest for each sample in descending order of cost. Each forest is encoded as an acyclic predecessor map $\mathcal{P}$ which assigns to each sample $\bm{x}_i \notin \mathcal{R}$ its predecessor $\mathcal{P}(\bm{x}_i)$ in the optimum path from $\mathcal{R}$, or a marker $nil$ when $\bm{x}_i \in \mathcal{R}$.

It is important to remark that the unsupervised OPF algorithm determines the number of clusters (OPTs) automatically, hence such information is not required beforehand, differently from other algorithms. Furthermore, the only hyperparameter that must be set is the search interval upper bound $k_{\max}$ for the proper neighborhood size $k^\star \in [1, k_{\max}]$.

\subsection{Supervised Optimum-Path Forest}
\label{ss.supervised_opf}

Differently from its unsupervised version, the supervised variant uses a fully-connected graph $\mathcal{G}^\prime=(\mathcal{D}, \mathcal{B})$ instead. Moreover, the closest samples from different classes are marked as prototypes, as aforementioned. Regarding the competition process, instead of using Equation~\ref{e.f_min}, the following smooth function is employed:

\begin{eqnarray}
    \label{e.f_max}
    f_{\max}(\langle \bm{x}_i \rangle) & = &
        \left\{ \begin{array}{ll} 
            0           & \mbox{if $\bm{x}_i \in \mathcal{R}$} \\
            \infty      & \mbox{otherwise.} \\
        \end{array}\right.\\ \nonumber
    f_{\max}(\pi_{\bm{x}_i} \cdot \langle\bm{x}_i, \bm{x}_j\rangle) & = &
        \max\left\{
            f_{\max}(\pi_{\bm{x}_i}), d(\bm{x}_i, \bm{x}_j)
        \right\}.
\end{eqnarray}

Further, the following cost map is used to partition the graph:

\begin{equation}
    \label{e.max_cost_map}
    \mathcal{S}(\bm{x}_i) = \min_{\bm{x}_i \in \mathcal{D}}
    \left\{
        \max \left\{
            \mathcal{S}(\bm{x}_i),
            d(\bm{x}_i, \bm{x}_j)
        \right\}
    \right\}.
\end{equation}

Such costs are used to initialize the algorithm before evaluating Equation~\ref{e.max_cost_map}, which is performed for every node in an ascending order of costs. After partitioning the graph, each prototype propagates its ground truth label to all samples in its OPT. Afterwards, prediction is performed by solving the same equation for new samples individually.

\section{Proposed Approach}
\label{s.proposed}

Although the unsupervised OPF was conceived for clustering purposes, such clusters possess a set of features that can also be employed to synthesize new samples, thus being suitable for oversampling. This section describes the procedure for a binary classification problem, which intends to oversample the class composed of the smallest number of features. Notwithstanding, the same approach can be easily applied to multiclass problems by individually repeating the procedure for each class to be oversampled.

\begin{figure}[h]
  \centerline{	
	\begin{tabular}{c}
      	\includegraphics[width=0.75\columnwidth]{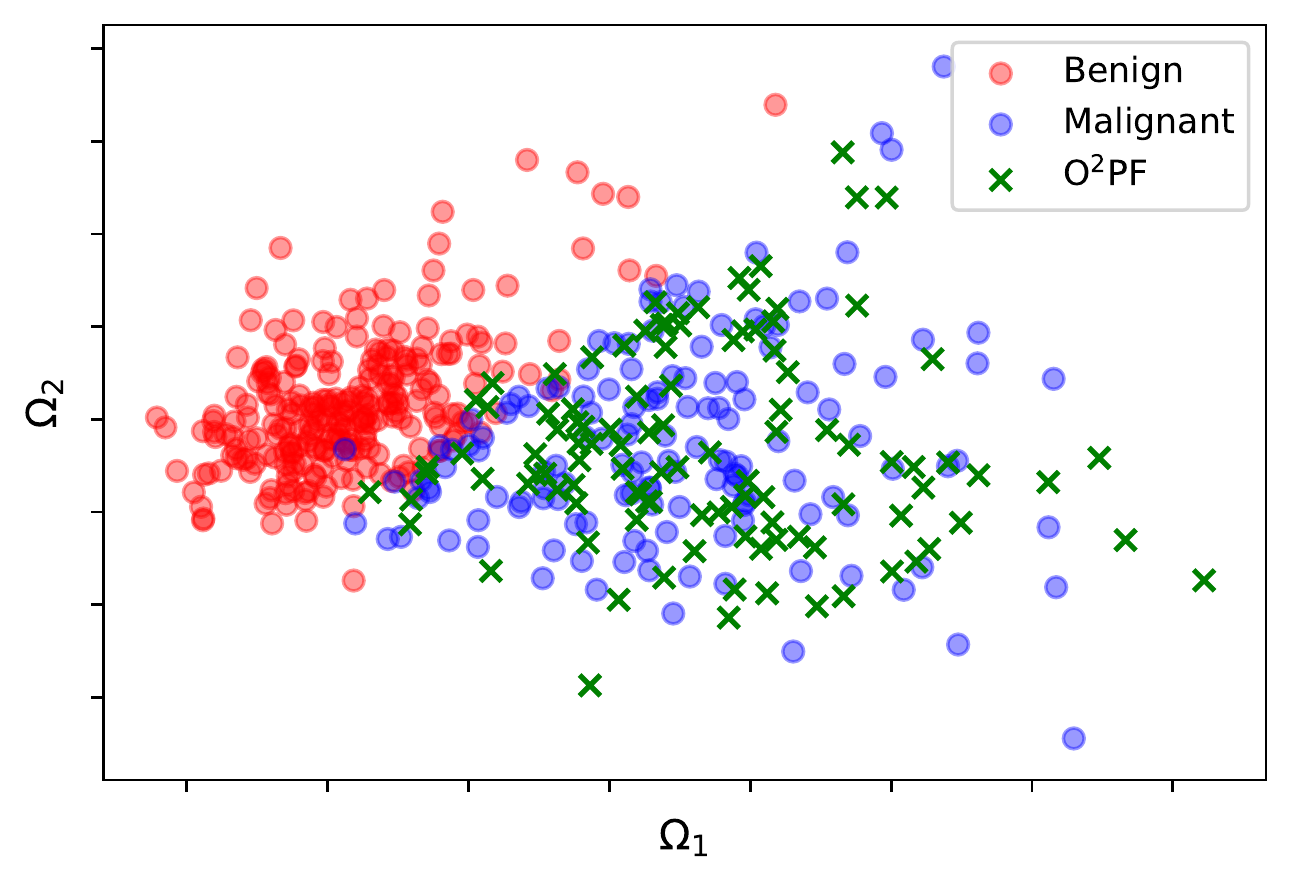}   \\
	\end{tabular}}
\caption{Oversampling using $\text{O}^2$PF. The image comprises WBCD Diagnostic II dataset samples and synthetic points from the minority class generated using $\text{O}^2$PF. Data were projected onto a $2$-dimensional space using the Principal Component Analysis algorithm for visualization purposes.}
\label{f.O2PF}
\end{figure}

The process of synthetic samples generation contemplates two main steps: (i)~creating plausible samples, i.e., synthetic elements with characteristics that are coherent with the selected class; and (ii)~introducing sample variability, avoiding making the classifier biased towards a subset of characteristics present in that class. To tackle such issues, the $\text{O}^2$PF first performs the clustering from minority class samples, which turns possible the extraction of common patterns intrinsic to the class, i.e., the samples' average position and variance. Further, the algorithm assumes that all features from a class follow a normal distribution. Thus, a new sample $\bm{z} \in \mathbb{R}^m$ can be generated by sampling such a distribution from some of the $q$ clusters found. Notice the number of synthetic samples generated by each cluster is proportional to the number of original samples compounding it, i.e., we are always doubling the number of samples from the minority class, although the user can set that percentage. The distribution is performed as follows:

\begin{equation}
    \label{e.synthetic}
    \bm{z} \sim \mathcal{N}(\bm{\mu}_q, \Sigma_q),
\end{equation}

\noindent
where $\bm{\mu}_q \in \mathbb{R}^m$ stands for the distribution mean, defined as the average feature vector from all samples within the $q$-th cluster. Moreover, ${\Sigma_q \in \mathbb{R}^{m \times m}}$ is the covariance matrix, computed as follows:

\begin{equation}
    \label{e.cov}
    \Sigma_q = \frac{1}{n_q - 1} (X_q - \bm{\mu}_q)(X_q - \bm{\mu}_q)^T,
\end{equation}

\noindent
given that $X_q \in \mathbb{R}^{m \times n_q}$ is a matrix formed by concatenating all $n_q$ cluster feature vectors. Figure~\ref{f.O2PF} illustrates the proposed approach behavior.

\section{Methodology}
\label{s.methodology}

This section describes the datasets used in the experiments. Further, the experimental setup is outlined.

\subsection{Datasets}
\label{ss.datasets}

Experiments were conducted over two sets of three databases each. The first set comprises three datasets for breast cancer detection, i.e., the Wisconsin Breast Cancer Database, which is composed of the datasets Prognostic, Diagnostic I, and Diagnostic II. The second set is composed of general-purpose datasets for medical issues. All of them are unbalanced, binary, and were obtained from the UCI repository~\cite{UCI:2019}. A brief description of each one follows below:

\begin{itemize}
  \item \textbf{Wisconsin Breast Cancer Database (WBCD) Prognostic}\footnote{All considered versions are available at \url{https://archive.ics.uci.edu/ml/datasets/Breast+Cancer+Wisconsin+(Prognostic)}.}~\cite{mangasarian1990cancer}\textbf{:} Regards predicting a sample as recurrent or non-recurrent type of cancer based on $32$ features. There are $198$ samples, being $151$ ($76.3\%$) non-recurrent and $47$ ($23.7\%$) recurrent;

  \item \textbf{WBCD Diagnostic I:} Consists of classifying a tumor as benignant or malignant based on $32$ features as well. There are $569$ instances, from which $357$ ($63.7\%$) are benign and $212$ ($37.3\%$) are malignant;

  \item \textbf{WBCD Diagnostic II:} Corresponds to labelling each of the $699$ samples as benign or malignant tumor. Each sample comprises $9$ features and each class contains $458$ ($65.5\%$) and $241$ ($34.5\%$) examples, respectively;

  \item \textbf{Diabetic Retinopathy Debrecen (DRD)}\footnote{Available at \url{https://archive.ics.uci.edu/ml/datasets/Diabetic+Retinopathy+Debrecen+Data+Set}.}~\cite{antal2014ensemble}\textbf{:} Regards predicting whether an image contains signs of diabetic retinopathy or not based on $19$ variables. The dataset contains $1,151$ samples, from which $611$ ($53.1\%$) are positive and $540$ ($46.9\%$) are negative;

  \item \textbf{Cervical Cancer (CC)}\footnote{Available at \url{https://archive.ics.uci.edu/ml/datasets/Cervical+cancer+\%28Risk+Factors\%29}.}~\cite{fernandes2017transfer}\textbf{:} For this task we predict the binary \textit{biopsy} variable based on $32$ features for $858$ samples. Differently from other datasets, all variables in this scenario are either integer or binary. Further, the dataset is significantly skewed, with $55$ $(6.4\%)$ positive and $803$ ($93.6\%$) negative samples;

  \item \textbf{Mammographic Mass (MM)}\footnote{Available at \url{https://archive.ics.uci.edu/ml/datasets/Mammographic+Mass}.}~\cite{elter2007prediction}\textbf{:} Concerns predicting if a mammographic mass is benign or malignant based on six features. The dataset contains $516$ ($53.7\%$) benign and $445$ ($46.3\%$) malignant samples, comprising $961$ examples.
\end{itemize}

\subsection{Experimental setup}
\label{ss.setup}

The experiments conducted in this paper considered pre-processing the data such that missing features were replaced by their corresponding mean in the training partition. All the features were normalized to have zero mean and unitary standard deviation. Further, the datasets were randomly divided into training, validation and testing sets, each containing $70\%$, $15\%$, and $15\%$ of the data, respectively\footnote{Such percentages were obtained empirically.}. 

\begin{figure}[h]
  \includegraphics[width=\linewidth]{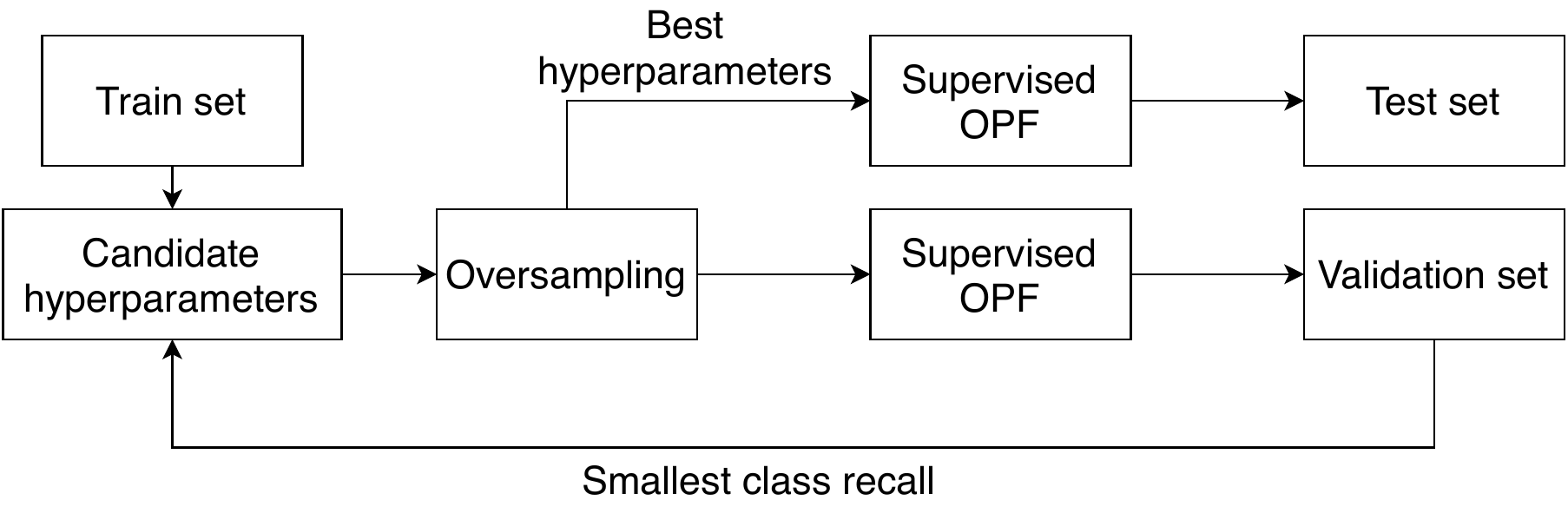}
  \caption{Experimental pipeline for each dataset partition.}
  \label{f.pipeline}
\end{figure}

The validation set was employed to fine-tune the oversampling method hyperparameters, i.e., finding the $k_{\max}$ for the unsupervised OPF and $\kappa$ for the other methods that maximize the minority class recall. Afterward, the augmented dataset is used to train the OPF classifier for further computing the results over the testing sets. Notice that such a procedure is repeated $20$ times for statistical analysis, and the results are compared through the Wilcoxon signed-rank test~\cite{Wilcoxon:45} with $0.05$ significance concerning recall values. Figure~\ref{f.pipeline} depicts such a pipeline. Implementation-wise, we rely on the supervised and unsupervised implementations provided by Opfython~\cite{rosa2020opfython}. Additionally, the source code was implemented using Python and is available on GitHub.\footnote{The source code is available online at \url{https://github.com/Leandropassosjr/o2pf}.}

\section{Experimental Results}
\label{s.experiments}

This section is divided into four main steps: (i) datasets augmented using the $\text{O}^2$PF are compared against the standard version, (ii) the proposed approach is compared against three baselines for oversampling considering three distinct versions of the Wisconsin Breast Cancer Database, i.e., Prognostic, Diagnostic I, and Diagnostic II, for the task of breast cancer detection. In step (iii), a similar experiment is conducted over three general-purpose medical issues datasets, and (iv) it provides a brief discussion concerning the optimization of the proposed method hyperparameter, i.e., the $k_{\max}$.
 
\subsection{$\text{O}^2$PF Data Augmentation Versus Standard Datasets}
\label{ss.standard}

This section presents the results obtained by the Optimum-Path Forest classifiers considering the datasets balanced through $\text{O}^2$PF oversampling, i.e., minority classes are augmented such that both classes present a similar number of samples, against the standard version of the datasets. Table~\ref{t.proposedxStandard} presents the recall considering each dataset. Values in bold denote the best results according to the Wilcoxon signed-rank test with $5\%$ of significance.

\begin{table}[!htb]
\caption{{Evaluation of the proposed $\text{O}^2$PF against the standard datasets versions.}}
\begin{center}
\renewcommand{\arraystretch}{1.5}
\setlength{\tabcolsep}{6pt}
\resizebox{\columnwidth}{!}{
\begin{tabular}{c|c|c|c|c|c|c|c}
\hhline{-|-|-|-|-|-|-|-|}
\hhline{-|-|-|-|-|-|-|-|}
\hhline{-|-|-|-|-|-|-|-|}
{\cellcolor[HTML]{EFEFEF}{\textbf{Ds. Version}}} & {\cellcolor[HTML]{EFEFEF}{\textbf{Recall}}} & {\cellcolor[HTML]{EFEFEF}{\textbf{Prognostic}}} & {\cellcolor[HTML]{EFEFEF}{\textbf{Diagnostic I}}} & {\cellcolor[HTML]{EFEFEF}{\textbf{Diagnostic II}}} & {\cellcolor[HTML]{EFEFEF}{\textbf{DRD}}} & {\cellcolor[HTML]{EFEFEF}{\textbf{CC}}} & {\cellcolor[HTML]{EFEFEF}{\textbf{MM}}}\\ \hline
\multirow{2}{*}{ORIGINAL} & Avg. & $0.4945$ & $\bm{0.9184}$ & $0.8903$ & $0.4922$ & $\bm{0.6320}$ & $\bm{0.6180}$\\
&  Std. & $\pm0.1872$ & $\pm0.0346$ & $\pm0.0550$ & $\pm0.2017$ & $\pm0.1002$ & $\pm0.0480$\\\hline\hline
\multirow{2}{*}{$\text{O}^2$PF} & Avg. & $\bm{0.5739}$ & $\bm{0.9311}$ & $\bm{0.9104}$ & $\bm{0.5834}$ & $\bm{0.6292}$ & $0.6086$\\
&  Std. & $\pm0.1368$ & $\pm0.0331$ & $\pm0.0452$ & $\pm0.2308$ & $\pm0.0924$ & $\pm0.0434$\\
\hhline{-|-|-|-|-|-|-|-|}
\hhline{-|-|-|-|-|-|-|-|}
\hhline{-|-|-|-|-|-|-|-|}
\end{tabular}}
\label{t.proposedxStandard}
\end{center}
\end{table}

In this context, one can observe $\text{O}^2$PF obtained the best results in five out of six datasets according to the Wilcoxon signed-rank test, obtaining the best results alone in three of them. 

\subsection{Results concerning the Breast Cancer datasets}
\label{ss.resultsBreast}

Table~\ref{t.results_Breast} presents the average recall, accuracy, F1-measure and best $k_{\max}$ considering $\text{O}^2$PF or $\kappa$ considering the other techniques, as well as their standard deviation. The results comprise a minority oversampling, thus providing balanced datasets. The proposed approach is compared against three baseline techniques, i.e., SMOTE, Borderline SMOTE, and ADASYN.

\begin{table}[!htb]
\caption{Results considering WBCD Breast Cancer datasets.}
\begin{center}
\renewcommand{\arraystretch}{1.5}
\setlength{\tabcolsep}{6pt}
\resizebox{\columnwidth}{!}{
\begin{tabular}{c|c|c|c|c|c}
\hhline{-|-|-|-|-|-|}
\hhline{-|-|-|-|-|-|}
\hhline{-|-|-|-|-|-|}
{\cellcolor[HTML]{EFEFEF}{\textbf{Dataset}}} & {\cellcolor[HTML]{EFEFEF}{\textbf{Statistics}}} & {\cellcolor[HTML]{EFEFEF}{\textbf{$\text{{O}}^2$PF}}} & {\cellcolor[HTML]{EFEFEF}{\textbf{SMOTE}}} & {\cellcolor[HTML]{EFEFEF}{\textbf{Borderline SMOTE}}} & {\cellcolor[HTML]{EFEFEF}{\textbf{ADASYN}}}\\ \hline
\multirow{4}{*}{Prognostic} & Recall & $\bm{0.5739\pm0.1368}$ & $\bm{0.5803\pm0.1253}$ & $\bm{0.5802\pm0.1343}$ & $\bm{0.6311\pm0.1283}$\\
&  Accuracy & $0.6317\pm0.0654$ & $0.6183\pm0.0532$ & $0.6317\pm0.0619$ & $0.6300\pm0.0547$\\
&  F1 & $0.4143\pm0.1310$ & $0.4112\pm0.1330$ & $0.4175\pm0.1266$ & $0.4366\pm0.1413$\\
&  Best $k$ & $37.2500\pm27.0867$ & $6.4000\pm1.5297$ & $6.0500\pm1.4654$ & $6.1500\pm1.4586$\\\hline\hline
\multirow{4}{*}{Diagnostic I} & Recall & $\bm{0.9311\pm0.0331}$ & $\bm{0.9368\pm0.0275}$ & $\bm{0.9347\pm0.0322}$ & $\bm{0.9422\pm0.0336}$\\
&  Accuracy & $0.9471\pm0.0170$ & $0.9500\pm0.0138$ & $0.9424\pm0.0178$ & $0.9448\pm0.0191$\\
&  F1 & $0.9285\pm0.0257$ & $0.9330\pm0.0198$ & $0.9231\pm0.0278$ & $0.9264\pm0.0285$\\
&  Best $k$ & $12.7500\pm12.6960$ & $5.4000\pm0.8602$ & $5.6500\pm1.2359$ & $6.0000\pm1.6432$\\\hline\hline
\multirow{4}{*}{Diagnostic II} & Recall & $\bm{0.9104\pm0.0452}$ & $\bm{0.9075\pm0.0541}$ & $0.8947\pm0.0586$ & $\bm{0.9108\pm0.0520}$\\
&  Accuracy & $0.9490\pm0.0176$ & $0.9476\pm0.0192$ & $0.9419\pm0.0217$ & $0.9490\pm0.0191$\\
&  F1 & $0.9195\pm0.0294$ & $0.9170\pm0.0331$ & $0.9080\pm0.0353$ & $0.9196\pm0.0315$\\
&  Best $k$ & $23.5000\pm29.6268$ & $6.4500\pm1.4992$ & $5.4000\pm0.8602$ & $6.8500\pm1.5580$\\
\hhline{-|-|-|-|-|-|}
\hhline{-|-|-|-|-|-|}
\hhline{-|-|-|-|-|-|}
\end{tabular}}
\label{t.results_Breast}
\end{center}
\end{table}

Results observed over WBCD Prognostic dataset show all techniques obtained similar statistical results considering the recall. Regarding the accuracy, one can observe the proposed approach obtained the highest average value, together with the borderline SMOTE. Such a result suggests samples generated by $\text{O}^2$PF fit better the class distribution, being less prone to false negatives. A similar behavior is observed over the Diagnostic I dataset. Concerning the Diagnostic II dataset, the proposed approach obtained the best results, together with ADASYN and SMOTE, outperforming Borderline SMOTE. Considering the accuracy, $\text{O}^2$PF obtained the highest averages.

Despite the good performance achieved by the oversampling approach using $\text{O}^2$PF, the baseline algorithms have shown statistical similarity in most of the cases. Sparse clusters with low density may be responsible for introducing new samples that are distant from the original distribution of the minority class, a situation that would be likely to produce outliers in the new resample dataset. Such behavior may also influence in a moderate recall, as observed in Tables~\ref{t.results_Breast}.


\subsection{General Purpose Medical Datasets Results}
\label{ss.resultsGeneral}

Table~\ref{t.results_general} presents the results obtained over the three general purpose medical datasets, i.e., Diabetic Retinopathy Debrecen, Cervical Cancer and Mammographic Mass.

\begin{table}[!htb]
\caption{Results considering General Purpose Datasets.}
\begin{center}
\renewcommand{\arraystretch}{1.5}
\setlength{\tabcolsep}{6pt}
\resizebox{\columnwidth}{!}{
\begin{tabular}{c|c|c|c|c|c}
\hhline{-|-|-|-|-|-|}
\hhline{-|-|-|-|-|-|}
\hhline{-|-|-|-|-|-|}
{\cellcolor[HTML]{EFEFEF}{\textbf{Dataset}}} & {\cellcolor[HTML]{EFEFEF}{\textbf{Statistics}}} & {\cellcolor[HTML]{EFEFEF}{\textbf{$\text{{O}}^2$PF}}} & {\cellcolor[HTML]{EFEFEF}{\textbf{SMOTE}}} & {\cellcolor[HTML]{EFEFEF}{\textbf{Borderline SMOTE}}} & {\cellcolor[HTML]{EFEFEF}{\textbf{ADASYN}}}\\ \hline
\multirow{4}{*}{DRD} & Recall & $\bm{0.6086\pm0.0434}$ & $\bm{0.5831\pm0.0476}$ & $\bm{0.5855\pm0.0498}$ & $0.5716\pm0.0487$\\
&  Accuracy & $0.5934\pm0.0327$ & $0.5983\pm0.0353$ & $0.5974\pm0.0343$ & $0.5957\pm0.0354$\\
&  F1 & $0.6092\pm0.0302$ & $0.5799\pm0.0465$ & $0.5803\pm0.0460$ & $0.5734\pm0.0469$\\
&  Best $k$ & $26.2500\pm26.9664$ & $6.8500\pm1.6515$ & $7.5000\pm1.7748$ & $5.0000\pm0.0000$\\\hline\hline
\multirow{4}{*}{CC} & Recall & $\bm{0.5834\pm0.2308}$ & $\bm{0.6287\pm0.1505}$ & $0.5791\pm0.1811$ & $\bm{0.6237\pm0.1560}$\\
&  Accuracy & $0.9388\pm0.0212$ & $0.9403\pm0.0157$ & $0.9403\pm0.0165$ & $0.9415\pm0.0138$\\
&  F1 & $0.5330\pm0.1548$ & $0.5673\pm0.0900$ & $0.5414\pm0.1109$ & $0.5675\pm0.0905$\\
&  Best $k$ & $13.2500\pm10.8714$ & $6.5500\pm1.4654$ & $6.5500\pm1.7168$ & $6.0000\pm1.5166$\\\hline\hline
\multirow{4}{*}{MM} & Recall & $\bm{0.6292\pm0.0924}$ & $\bm{0.6757\pm0.0739}$ & $\bm{0.6876\pm0.0608}$ & $\bm{0.6614\pm0.0706}$\\
&  Accuracy & $0.6769\pm0.0531$ & $0.6738\pm0.0503$ & $0.6852\pm0.0426$ & $0.6876\pm0.0396$\\
&  F1 & $0.6439\pm0.0639$ & $0.6591\pm0.0545$ & $0.6710\pm0.0479$ & $0.6631\pm0.0567$\\
&  Best $k$ & $48.0000\pm34.2199$ & $7.1000\pm1.4799$ & $7.2500\pm1.5772$ & $6.1500\pm1.5898$\\
\hhline{-|-|-|-|-|-|}
\hhline{-|-|-|-|-|-|}
\hhline{-|-|-|-|-|-|}
\end{tabular}}
\label{t.results_general}
\end{center}
\end{table}

Considering Diabetic Retinopathy Debrecen, the proposed approach outperformed the average recall overall techniques, although SMOTE and Borderline SMOTE achieved similar results. Similar behavior is observed over the F1 metric. On the other hand, $\text{{O}}^2$PF, SMOTE and ADASYN obtained the best results regarding the Cervical Cancer dataset. Since both Borderline SMOTE and ADASYN are variants of SMOTE, they are expected to perform differently over different scenarios. However, as observed in most of the experiments, they generally are outperformed by SMOTE technique itself, considering the average values. Regarding Mammographic Mass datasets, all techniques performed in a very much alike fashion, obtaining similar results. 

One can notice that all three datasets present a challenging task since no technique reached a $0.7$ recall. Unusual behavior is observed over the Cervical Cancer dataset, whose all techniques obtained an approximate accuracy of $94\%$, despite the recall below $0.63$. Such behavior may suggest samples from the minority class are distributed among a subcluster from majority class, therefore providing high accuracy despite the low recall.

\subsection{$\text{O}^2$PF Hyperparameter Selection}
\label{ss.hyperparameter}

\begin{figure}[!ht]
  \centerline{	
	\begin{tabular}{cc}
		\includegraphics[width=0.5\columnwidth]{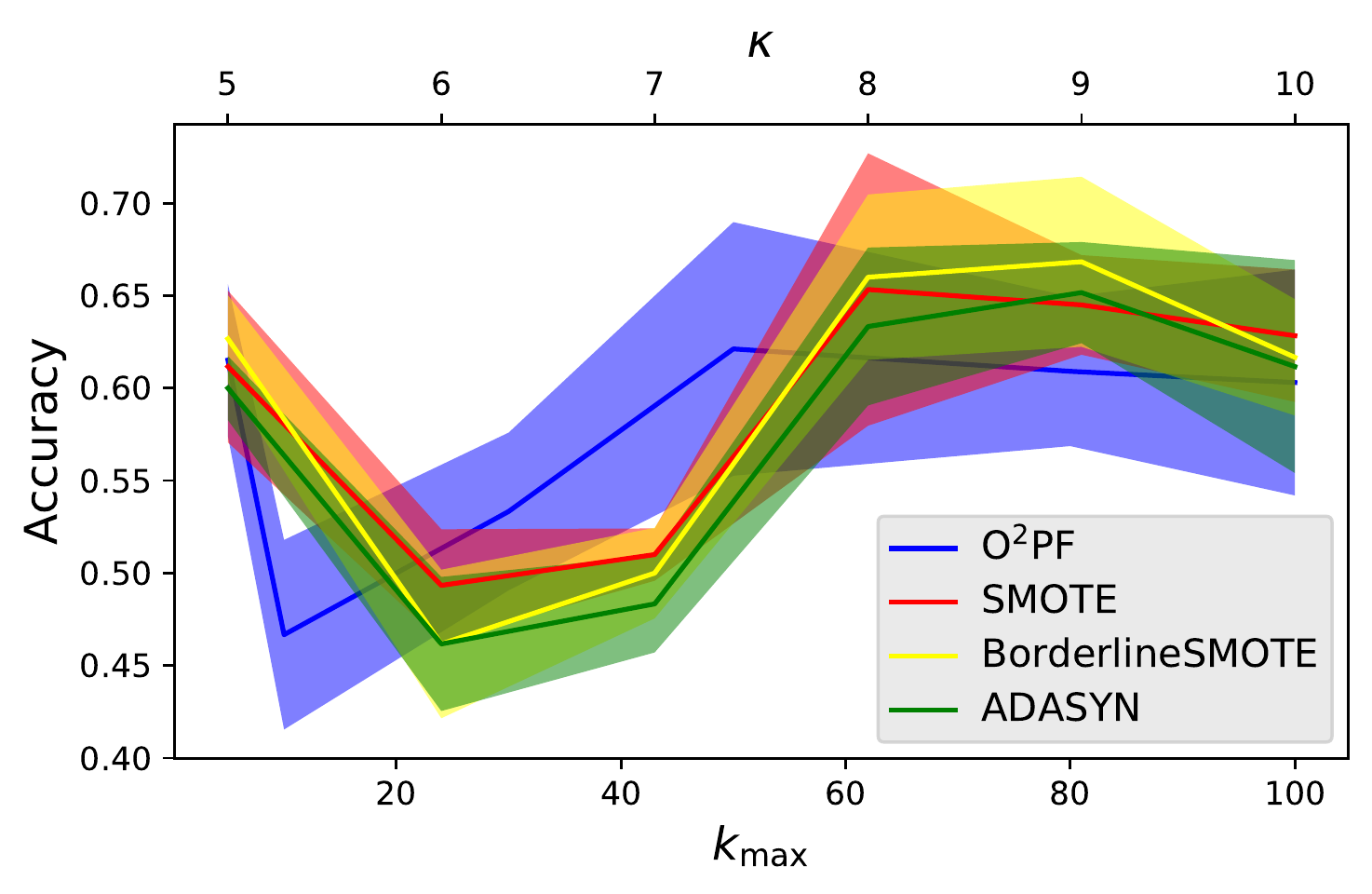}&
		\includegraphics[width=0.5\columnwidth]{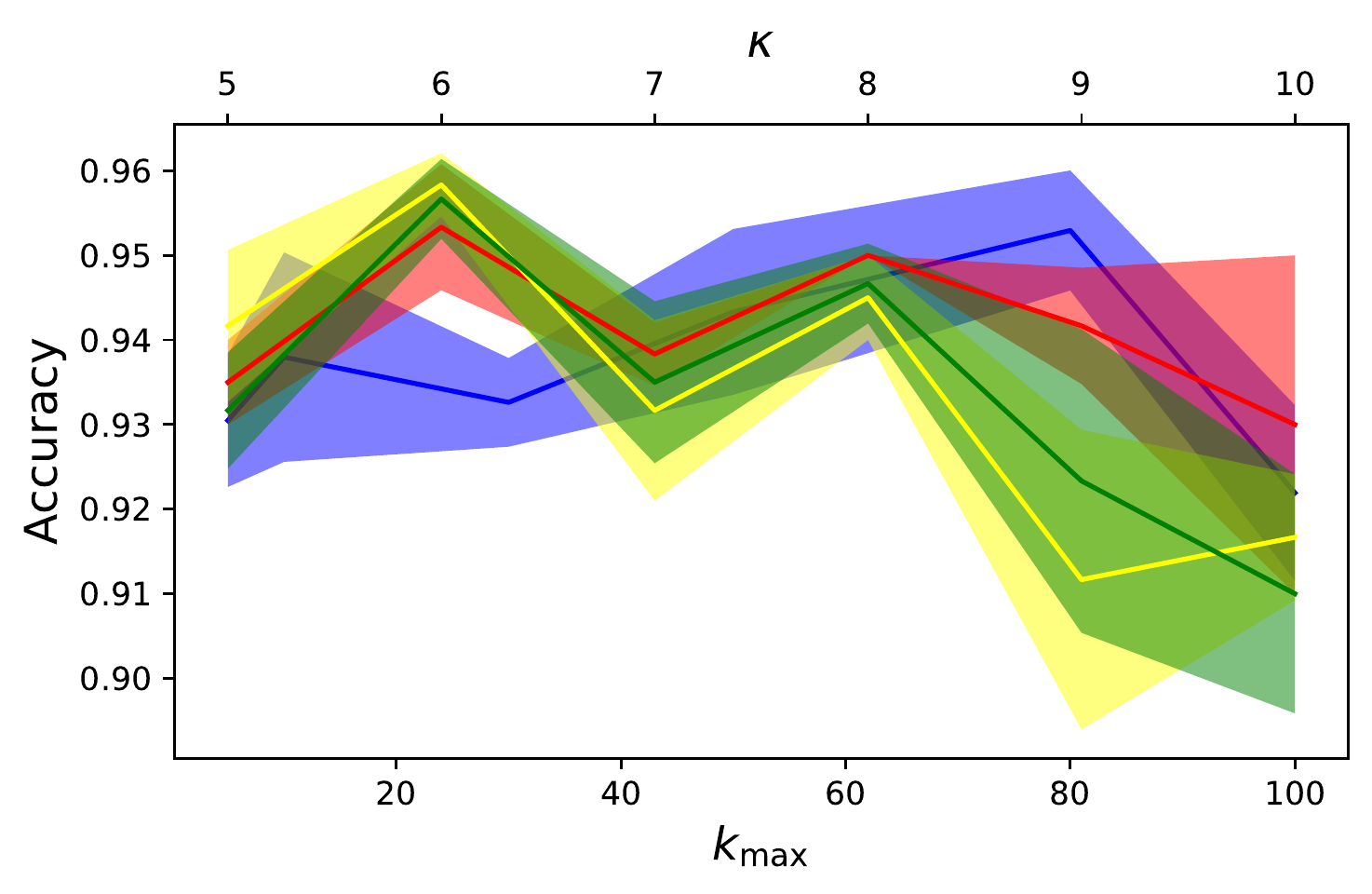}\\
      (a) &  (b) 
	\end{tabular}}
  \centerline{	
	\begin{tabular}{c}
		\includegraphics[width=0.5\columnwidth]{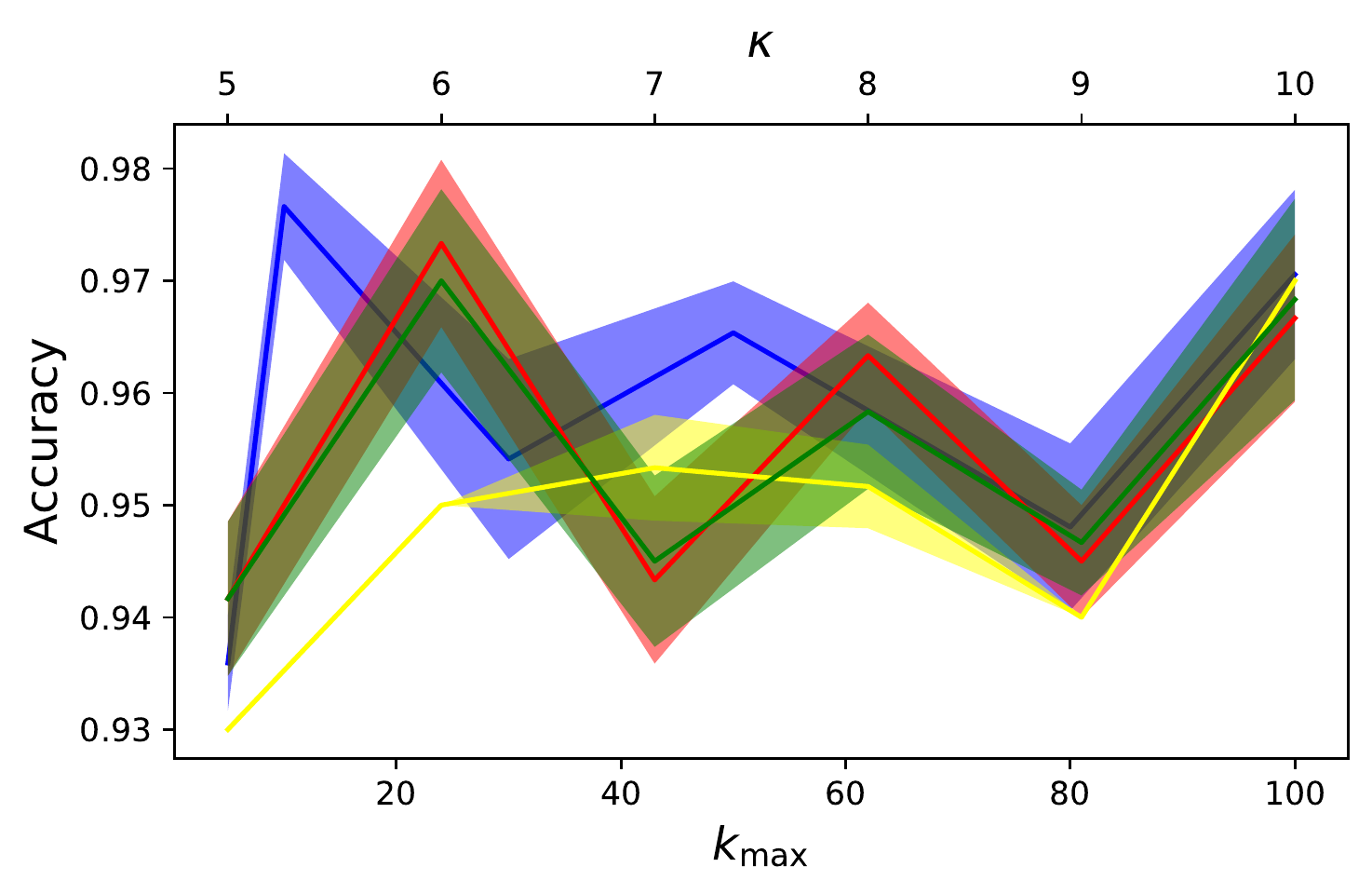}\\
      (c) 
	\end{tabular}}
\caption{Grid searching for a proper selection of the techniques' hyperparameter over the validation for WBCD (a) Prognostic, (b) Diagnostic I, and (c) Diagnostic II datasets.}
\label{f.plotProg}
\end{figure}

$\text{O}^2$PF requires a proper selection of a single hyperparameter, the $k_{max}$, which is employed in the clustering process. Such a hyperparameter, however, is way less sensitive when compared to a proper selection of the best $k$, performed by the other techniques. Figure~\ref{f.plotProg} depicts a grid search considering a proper selection of those hyperparameters for each technique. Notice the central line describes the average value over the validation dataset, while the broader area describes the standard deviation. Notice even though $\text{O}^2$PF considers a very wider interval, i.e., ranging from $[5,100]$, most of the time, the results outperform the other techniques, which assume a shorter interval between $[5,10]$.

\section{Conclusion}
\label{s.conclusion}
This paper presented an oversampling approach based on the unsupervised Optimum-Path Forest Algorithm. The proposed $\text{O}^2$PF showed to be capable of handling the class imbalance problem through a simple and effective procedure that generates new synthetic samples based on the normal distribution of the feature vectors inside each cluster. The experiments performed in three datasets of breast cancer, apart from three complementary medical issue datasets, showed that the $\text{O}^2$PF approach demonstrated similar or superior results when compared to the baseline methods already proposed in the literature. 

Notwithstanding, the effectiveness of the proposed approach may still suffer in synthesizing new samples based on low-density clusters, a situation that may introduce noise samples in the training set and, consequently, affect creating of the prediction model. Future studies will be conducted to overcome the influence of sparse clusters with low density in the process of synthesizing new outliers. Moreover, experiments with a multiclass problem will also be performed in subsequent investigations.

\bibliographystyle{IEEEtran}
\bibliography{references}

\end{document}